%% file: example.tex
\documentclass{article}

\usepackage[final]{corl_2020} 

\usepackage{graphics} 
\usepackage{epsfig} 
\usepackage{mathptmx} 
\usepackage{times} 
\usepackage{amsmath} 
\usepackage{amssymb}  

\usepackage{wrapfig}
\usepackage{lipsum}
\usepackage{booktabs}
\newcommand{\ra}[1]{\renewcommand{\arraystretch}{#1}}
\usepackage{multirow, makecell}
\usepackage[ ruled, vlined]{algorithm2e}
\usepackage [english]{babel}
\usepackage [autostyle, english = american]{csquotes}
\usepackage{subcaption}
\usepackage{dblfloatfix}
\usepackage{hyperref}
\usepackage{footmisc}
\hypersetup{
    colorlinks=true,
    linkcolor=black,
    filecolor=blue,      
    urlcolor=blue
}

\usepackage[toc,page]{appendix}
\usepackage{diagbox}

\MakeOuterQuote{"}
\newcommand{\etal}{\textit{et al}.}
\usepackage{bibspacing}
\title{Learning hierarchical relationships for object-goal navigation}

\author{\textnormal{Yiding Qiu} \thanks{Equal contribution}\\
UC San Diego\\
{\tt\small yiqiu@eng.ucsd.edu}
\and
Anwesan Pal \footnotemark[1]\\
UC San Diego\\
{\tt\small a2pal@eng.ucsd.edu}
\and
Henrik I. Christensen\\
UC San Diego\\
{\tt\small hichristensen@eng.ucsd.edu}
}

\begin{document}
\maketitle

\begin{abstract}
Direct search for objects as part of navigation poses a challenge for small items. Utilizing context in the form of object-object relationships enable hierarchical search for targets efficiently. Most of the current approaches tend to directly incorporate sensory input into a reward-based learning approach, without learning about object relationships in the natural environment, and thus generalize poorly across domains. We present Memory-utilized Joint hierarchical Object Learning for Navigation in Indoor Rooms (MJOLNIR), a target-driven navigation algorithm, which considers the inherent relationship between target objects, and the more salient contextual objects occurring in its surrounding. Extensive experiments conducted across multiple environment settings show an $82.9\%$ and $93.5\%$ gain over existing state-of-the-art navigation methods in terms of the success rate (SR), and success weighted by path length (SPL), respectively. We also show that our model learns to converge much faster than other algorithms, without suffering from the well-known overfitting problem. Additional details regarding the supplementary material and code are available at \url{https://sites.google.com/eng.ucsd.edu/mjolnir}.
\end{abstract}



\input{files/introduction.tex}
\input{files/relatedwork.tex}
\input{files/methodology.tex}
\input{files/experiments.tex}
\input{files/conclusion.tex}

\acknowledgments{The authors would like to thank Army Research Laboratory (ARL) W911NF-10-2-0016 Distributed and Collaborative Intelligent Systems and Technology (DCIST) Collaborative Technology Alliance for supporting this research.}


\bibliography{example}  
\appendix
\input{files/appendix}
\end{document}

%% file: files/introduction.tex
\section{Introduction} \label{sect:intro}

\begin{figure}[t]
    \centering
    \includegraphics[width=0.7\linewidth]{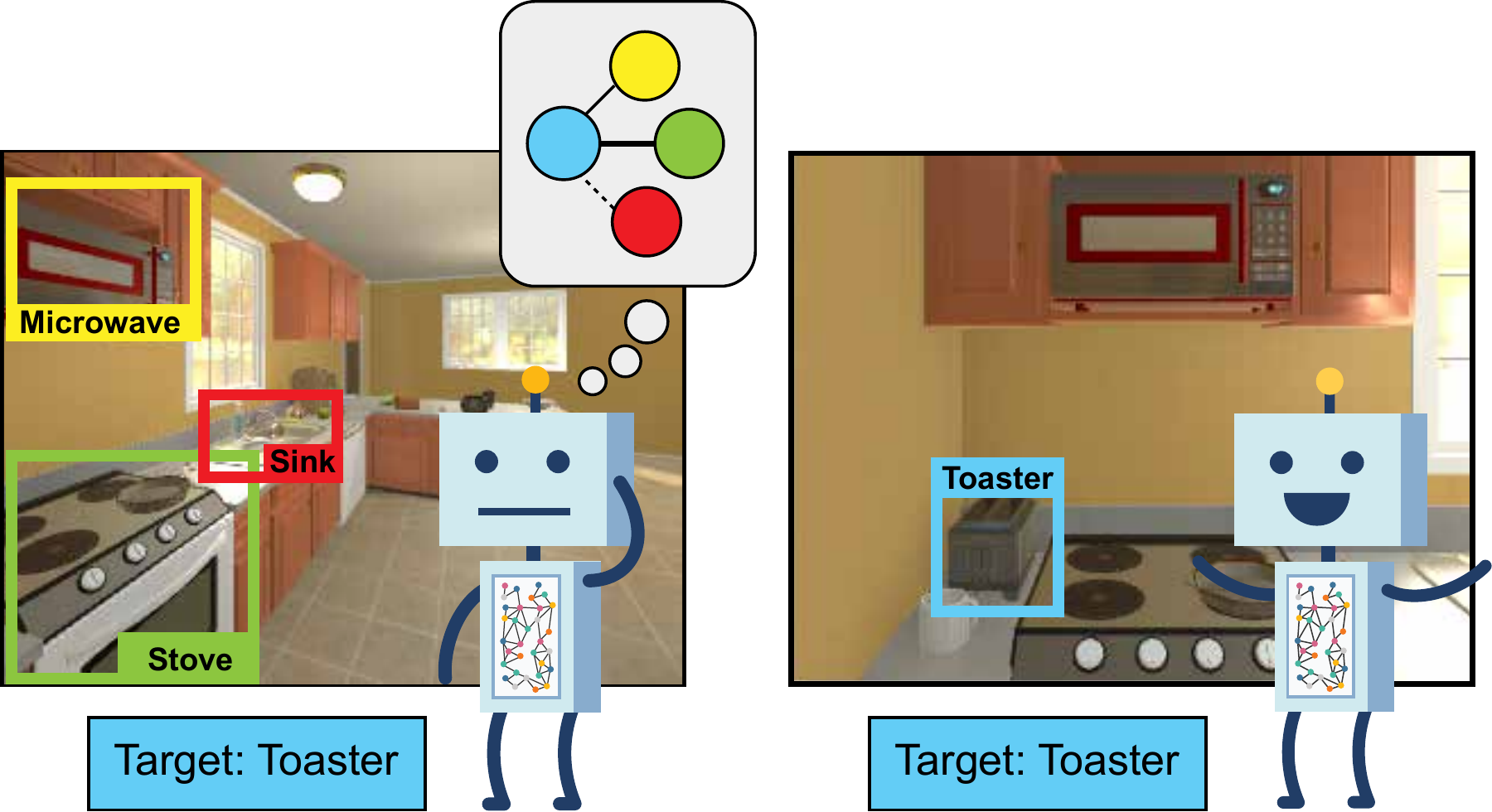}
    \caption{Illustration of the parent-target relationship. Upon seeing a set of parent objects (left image), the agent learns to associate the correct parent ($\mathtt{Stove}$ or $\mathtt{Microwave}$, here) to the target object ($\mathtt{Toaster}$, here) from the knowledge graph, and tries to search in its neighborhood to successfully locate the target object (right image). [Best when viewed in color]}
    \label{fig:intro}
\end{figure}

Human beings are able to perform complex tasks such as object-goal navigation efficiently, with implicit memorization of the relationships between the different objects. For example, to navigate to a target such as a $\mathtt{toaster}$ in the kitchen, a natural thing to do is to start by looking around a set of larger candidate objects which are likely to be nearby, such as a $\mathtt{stove}$ or $\mathtt{microwave}$. Unfortunately, this type of \textit{hierarchical relationship} is rarely used in robot navigation. As a result, the agent usually fails to develop any intuition about the goal location when they are further away. In this work, we refer to the object goal as $\mathtt{target}$, and those larger candidates which have either a spatial and semantic relationship with the target as $\mathtt{parent}$. This is depicted in Figure \ref{fig:intro}. Two main challenges are addressed here - (i) Correctly associating the robot's current observation with some prior intuition about object relationships into the model. (ii) Efficiently utilizing this hierarchical relationship for the visual navigation problem. 

Existing research in this area tends to aggregate sensory input into a meaningful state, before sending it to a Reinforcement Learning (RL) framework, with the expectation that the robot can implicitly learn the navigation problem through recursive trials. Zhu \etal \cite{zhu2017target} proposed to solve this problem by finding the similarity between current observation and the target observation through a trained Siamese Network \cite{koch2015siamese}. The work of Wortsman \etal \cite{Wortsman_2019_CVPR} incorporated a meta-learning approach where the agent learns a self-supervised interaction loss during inference to avoid collision. However, none of these methods use any prior information or semantic context. Sax \etal \cite{sax20a} highlight that a set of mid-level visual priors such as depth and edge information, surface normals, keypoints, etc. can be useful for the navigation task. However, a different encoder is required for learning each of these representations, and thus, the model does not scale very well. Moreover, the learnt features are expected to vary with the test setting, thereby limiting the scope of the method in unseen dissimilar environments. Instead, using the knowledge about object relationships is more robust since they are domain-independent. Two approaches which are similar to ours are Yang \etal \cite{yang2018visual}, and Druon \etal \cite{8963758}. In \cite{yang2018visual}, semantic priors in the form of a knowledge graph are used to capture object-object connectivity, but this connectivity is defined only in terms of their spatial proximity, not the inter-object dependency. For instance, an object such as $\mathtt{pillow}$ in the bedroom might be visible next to both a $\mathtt{bed}$ and an $\mathtt{alarm clock}$. Yet, we know from experience that to find a pillow, one should always start by looking around the bed. This type of common-sense knowledge is missing from their method. In contrast, our proposed approach assigns a sub-goal reward to force the agent to learn this key hierarchical relationship.
Also, important information pertaining to an object's spatial location in the scene is missing as they learn a single context for the whole scene. Instead, we utilize an object detector to capture this information explicitly. Recently, \cite{8963758} introduced the concept of a \textit{context grid} where they modeled the spatial (through an object detector) and language (through a word-embedding) similarity between the target and other objects as a $16\times 16$ grid. However, this relationship does not update during the training stage, thereby making the model less adaptable. Additionally, their action space is much bigger, thus simplifying the navigation task.


 A major issue with RL algorithms in a real-world setting is efficient modeling of the continuous high-dimensional state space in the agent's surrounding \cite{dulac2019challenges}. While implementing other algorithms \cite{zhu2017target, yang2018visual}, it was observed that even though the convergence during training stage was fast, they gave a relatively poor performance during testing. A possible reason for this is that due to their non-representative state space, the agent is perhaps simply learning to memorize the training setting after a certain number of episodes rather than understanding the underlying object relationships, and thus it fails to generalize to a novel scene. The adaptive gradient of Wortsman \etal \cite{Wortsman_2019_CVPR} alleviates this problem to some extent. Nevertheless, it doesn't remove it completely since it does not consider any prior memory into account. In contrast, our representative state leverages the correct balance of prior context and current observation to provide sufficient information about the surrounding during training, while simultaneously being abstract enough to generalize to a different room layout during testing.

\textbf{Contributions} We propose a hierarchical object relationship learning approach to the object-goal navigation problem by understanding the role of semantic context. This is done through the proposed novel \textit{Context Vector} as a node embedding in the graph convolutional neural network. We also emphasize the role object detection plays as opposed to a scene representation from traditional image classification networks. Finally, we incorporate the mentioned parent-target object relationship through a new reward shaping mechanism.

The remainder of the paper is organized as follows. In Section \ref{sect:relwrk}, we discuss related work, followed by the task definition in Section \ref{subsect: task_def}. The core ideas behind our approach are described in Section \ref{sect:method}. In Section \ref{sect:exp_res} we discuss the dataset used and the overall experimental design and results, before we summarize the work and outline future challenges in Section \ref{sect:conc}.

%% file: files/relatedwork.tex
\section{Related Work} \label{sect:relwrk}
\textbf{Map vs map-less navigation approaches} - Traditional approaches in visual navigation involved formalizing this as an obstacle avoidance problem, where the agent learns to navigate in its environment through a collision-free trajectory. This is done either in the form of offline maps \cite{borenstein1991vector, kim1999symbolic, 8968547}, or online maps \cite{davison2003real, sim2006autonomous, tomono2006map, wooden2006guide} generated through Simultaneous Localization and Mapping (SLAM) techniques. Given this map as input, the typical approach was to employ a path planning algorithm such as A* \cite{hart1968formal} or RRT* \cite{karaman2011sampling} to generate a collision-free trajectory to the goal. The limitation of these algorithms is that it might not be possible to have a pre-computed map of the environment. Generating a rich semantic map online is also a non-trivial task. With the advent of deep learning, the focus has instead shifted towards methods which are map-less \cite{chen2015deepdriving, gupta2017cognitive, giusti2015machine, linegar2016made, saeedi2006vision}, meaning that the representations can be learnt over time through interactions. However, most of these algorithms are not suited for finding specific target objects in a previously unseen environment. In contrast, our approach solves the target-driven navigation problem entirely using only visual inputs without the need of a pre-computed map.

\textbf{Point-goal navigation vs object-goal navigation} - Point-goal navigation \cite{anderson2018evaluation} refers to the problem where an agent starts from a randomly chosen pose and learns to navigate to a specific target point, usually specified in terms of 2D/3D coordinates relative to the agent. Lately, there has been some research \cite{mishkin2019benchmarking, kojima2019learn} in this area. However, it is still an ill-defined problem in a realistic setting, thereby making comparison difficult \cite{savva2019habitat}. Object-goal navigation refers to the problem where the agent instead learns to navigate to a specified target object while successfully avoiding obstacles. These tasks usually require some prior knowledge about the environment which can be useful for navigation \cite{zhu2017target, Wortsman_2019_CVPR, yang2018visual, 8963758}. Our approach falls in this category, but involves learning a robust contextual object-object relationship.

\textbf{Role of learning semantic context} - Learning semantic context of the surrounding world is an important research topic in the computer vision and robotics community. However, most of the existing work \cite{torralbacontext, hoiem2005geometric, rabinovich2007objects, mottaghi2014role, shrivastava2016contextual, marszalek2009actions} surrounds static settings such as object detection, semantic segmentation, activity recognition etc. Recently, object-object relationship modeling has been studied for tasks such as image retrieval \cite{johnson2015image} using scene graphs, visual relation detection \cite{zhang2017visual}, visual question-answering \cite{johnson2017clevr, marino2019ok}, place categorization \cite{8968108}, and driver saliency prediction \cite{Pal_2020_CVPR}. In this paper, we propose an algorithm which successfully learns to exploit hierarchical object relationships for object-goal visual navigation. 

\textbf{Reward shaping for policy networks} - Reward shaping is a method in reinforcement learning for engineering a reward function in order to provide more frequent feedback on appropriate behaviors \cite{Wiewiora2010}. In general, defining intermediate goals or sub-rewards for an interaction-based learning algorithm is non-trivial, since the environment model is not always known. However, a divide-and-conquer approach is often imperative to exploit the latent structure of a task to enable efficient policy learning \cite{tang-etal-2018-subgoal, bakker2004hierarchical, goel2003subgoal}. Specifically, for the object-goal visual navigation problem, it is important to learn the inherent "parent-target" object relationship for providing a meaningful feedback to the end-to-end training. For the policy network, we use the Asynchronous Advantage Actor-Critic (A3C) \cite{mnih2016asynchronous} algorithm to sample the action and the value at each step as per the approach of other models \cite{zhu2017target, Wortsman_2019_CVPR, yang2018visual}.

%% file: files/methodology.tex
\section{Task Definition} \label{subsect: task_def}

The goal of the object-goal navigation problem is to find a target object, defined through a set $T= \{t_1, \hdots, t_N\}$, in a given environment. The problem is defined purely from a vision perspective, and therefore any information about the environment in the form of a semantic or a topological map is not provided. The agent is spawned at a random location in the environment at the beginning of an episode. The input to the model is current observation in the form of RGB images, and the target object's word-embedding, rolled over each time-step. Using this, the agent has to sample an action $a$ from its trained policy given a set of actions $A$, where $a \in A=\{\mathtt{MoveAhead}$, $\mathtt{RotateLeft}$, $\mathtt{RotateRight}$, $\mathtt{LookUp}$, $\mathtt{LookDown}$ and $\mathtt{Done}\}$. The $\mathtt{MoveAhead}$ action takes the agent forward by $0.25$ meters, while the $\mathtt{RotateLeft}$ and $\mathtt{RotateRight}$ actions rotate it by $45$ degrees. Finally, the $\mathtt{Look}$ action tilts the camera up/down by $30$ degrees. An episode is considered a "success", if the target object is $\mathtt{visible}$, meaning the agent can detect it in the current frame, and is within a distance of $1.5$ meters from it. When the "$\mathtt{Done}$" action is sampled by the agent, the episode ends and the model checks if this criterion is met.


\section{Memory-utilized Joint hierarchical Object Learning for Navigation in Indoor Rooms (MJOLNIR)} \label{sect:method}

\textbf{Parent-Target object relationship} - In addition to the target objects, we introduce a new set of object classes, defined by the set $P= \{p_1,\hdots,p_M\}$. These "parent objects" consist of the larger objects present in a room, which also happen to be spatially/semantically related to the target object. For example, $\mathtt{Counter Top}$ is a parent object in the $\mathtt{Kitchen}$ and $\mathtt{Bathroom}$ scenes, while $\mathtt{Shelf}$ is a parent object in $\mathtt{Living\text{ }room}$ and $\mathtt{Bedroom}$. The set of parent objects, $P$, are manually picked for each room based on the strong correspondences they have with the target object list, $T$, in the knowledge graph (explained below). The aim of the navigation agent is to start by exploring the area around $p \in P$, eventually leading to the target object $t_i \in T$.

\begin{figure}[t]
    \centering
    \includegraphics[width=1\linewidth]{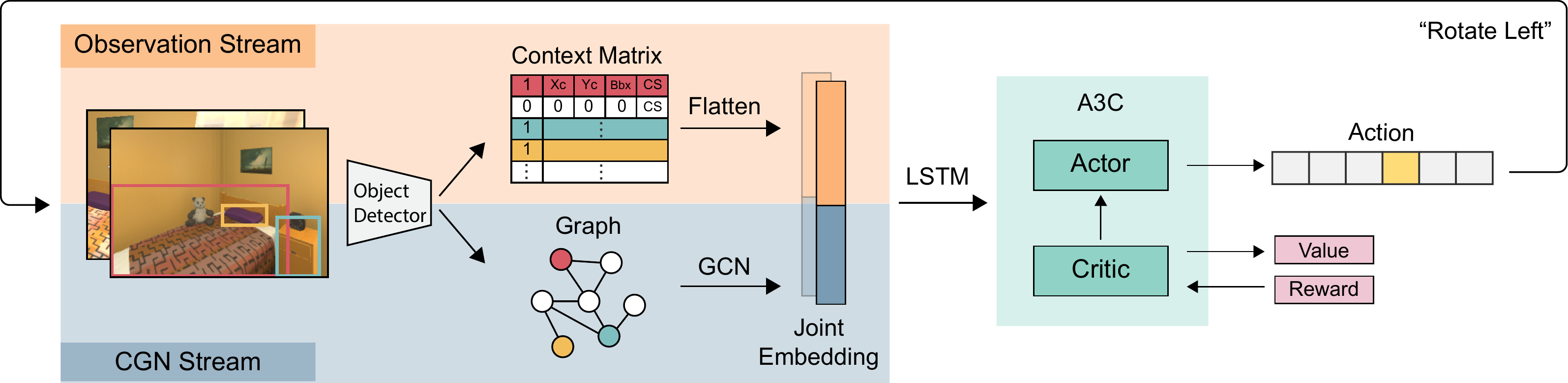}
    \caption{The entire MJOLNIR architecture. In the Observation stream, we use the ground-truth object detector to construct the context vector for all objects in our environment. This observation vector (orange) is concatenated with the graph embedding (blue) from the CGN stream to form a joint embedding. This is then sent to an LSTM cell and finally fed to the A3C model. [Best when viewed in color]}
    \label{fig:full_arch}
\end{figure}

\textbf{Construction of Knowledge Graph and the Context vector} \label{subsect:gcn} - Similar to \cite{yang2018visual}, our knowledge graph is also constructed using the objects and relationships extracted from the image-captions of the Visual Genome (VG) dataset \cite{krishna2017visual}. However, by pruning a lot of the object (for instance, "$\mathtt{armchair}$" vs "$\mathtt{arm\text{ }chairs}$") and relationship (for instance, "$\mathtt{near}$" vs "$\mathtt{next\text{ }to}$") aliases, we were able to build a cleaner adjacency matrix for the graph convolution network, containing strong relationship correspondences between those VG objects, which also appear in our experimental setting.

In addition to the newly constructed graph, we also introduce a novel $context$ $vector$ $\mathbf{c}_j$ for each object $o_j \in O$, where $O$ is the list of all the $101$ objects in our environment. This $5$-D vector gives information regarding the state of $o_j$ in the current frame, and can be represented as $\mathbf{c}_j = [b,x_c,y_c,bbox,CS]^T$. The first element, $b$, is a binary vector specifying whether $o_j$ can be detected in the current frame. The next two elements, ($x_c$, $y_c$), and $bbox$ correspond to the center ($x$, $y$) coordinates of the bounding box of $o_j$, and its covered area, both normalized with respect to the image size. Finally, $CS$ is a number giving the cosine similarity between the respective word embeddings of $o_j$, and the target object $t \in T$. This is expressed as:
\begin{equation*}
    CS(\mathbf{g}_{o_j},\mathbf{g}_t) = \frac{\mathbf{g}_{o_j}.\mathbf{g}_t}{||\mathbf{g}_{o_j}||.||\mathbf{g}_t||}
\end{equation*}
where $\textbf{g}$ denotes the word embeddings in the form of GloVe vectors \cite{pennington2014glove}.
\begin{figure}[t]
    \centering
    \includegraphics[width=1\linewidth]{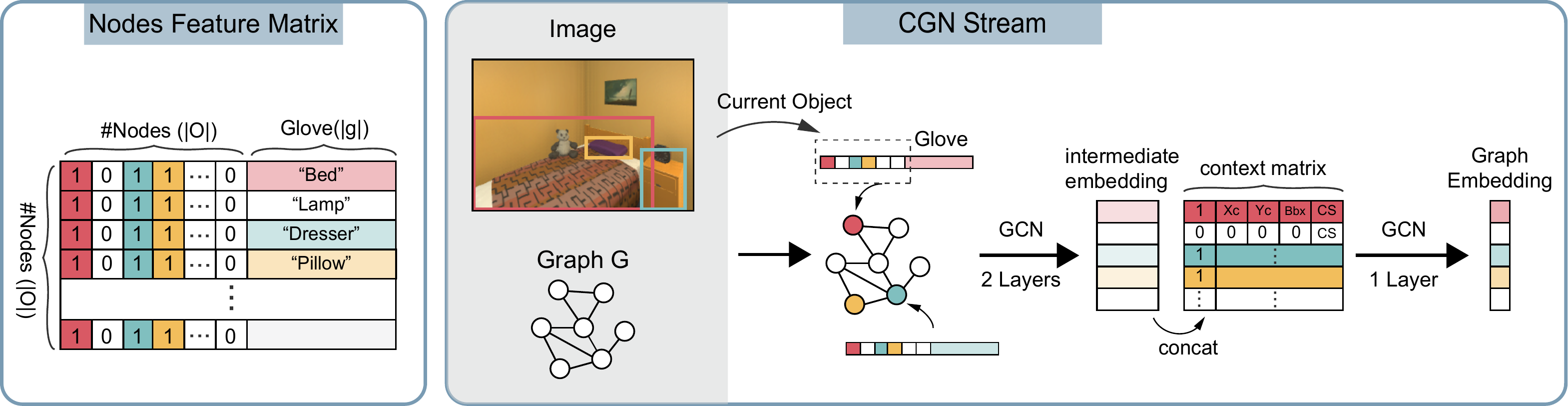}
    \caption{The novel CGN architecture. The input node feature constitutes the current detected object list and the node object's glove embedding. This is passed through two layers of GCN. This intermediate embedding is concatenated with the context vector and another layer of GCN is trained on it. [Best when viewed in color]}
    \label{fig:gcn_arch}
\end{figure}

\textbf{End-to-end model} - In this section, we talk about our entire network, shown in Figure \ref{fig:full_arch}. We adopt a two-stream network approach, consisting of (i) the Observation stream, which encodes the agent's current observation in the environment, and (ii) the Contextualized Graph Network (CGN) stream, which embeds the prior memory obtained through a knowledge graph $G = (V,E)$.  

For the Observation stream, we try two variants - (i) We use the ResNet-18 \cite{He_2016_CVPR} $\mathtt{conv}$ features obtained from the current frame to give a holistic representation of the scene. This feature map is then combined with the target object word embedding using point-wise convolution, and flattened to obtain the observation vector. (ii) We replaced the $\mathtt{conv}$ features with the $5$-D $context$ $vector$ (described above) for every object in our environment. The resulting $context$ $matrix \in R^{|O|\times 5}$ is then flattened and forms the observation vector.

Even though the input knowledge graph, $G$, provides a strong initial guidance to the agent, this information, by itself, can be insufficient due to the domain difference between VG \cite{krishna2017visual} and our navigation environment. The CGN stream is used to diminish this gap. We use a graph convolution network (gcn) \cite{kipf2016semi} to learn the node embeddings. Given $G$, we design the node feature vector $X \in R^{|O|+|\textbf{g}|}$ which is the concatenation of the output of an object detector on the current frame, specified by the $|O| = 101$-dimensional vector having $1$ for the current frame objects, and $0$ for others, along with each node object's word-embedding. "$101$" is the length of the exhaustive list of object types in our simulator, which is analogous to the number of object classes in a trained object detector. This is different from the $1000$-dimensional class probability used by Yang \etal \cite{yang2018visual}. The reason for this change is two-fold - (i) the probability vector obtained from ResNet-18 \cite{He_2016_CVPR} is pre-trained on the $1000$ classes of ImageNet \cite{deng2009imagenet}, which are quite different from the object list present in our environment, and (ii) since the pre-trained network primarily learns to classify a single object, in the multi-object setting, it is more likely to generate noisy labels\footnote{An illustration of this is shown in the supplementary videos in Appendix A}. The combined input node features are passed through two layers of gcn, to generate the intermediate embedding. This new feature is then concatenated with the $context$ $vector$, and then fed to another layer of gcn to generate the final graph node embedding. The concatenation of the observation vector and the graph embedding results in the joint embedding (shown in Figure \ref{fig:full_arch}), which is the input to the A3C model. The CGN stream is detailed in Figure \ref{fig:gcn_arch}.

To separately highlight the contribution of the changes made to each of the streams, we present two variants of our algorithm. MJOLNIR-r uses ResNet and word embedding in the current observation stream, while using CGN as the graph stream. MJOLNIR-o uses the flattened \textit{context matrix} as the observation vector, along with our CGN stream.

\textbf{Reward} \label{subsect:rew_shap} - We tune our reward function to correctly learn to utilize the parent-target relationship for the navigation task. The agent in our model receives a "partial reward", $R_p,$ when a parent object $p \in P$ is $\mathtt{visible}$. This is given by $R_p=R_t * Pr(t|p) * k$, where $R_t$ is the target reward and $k \in (0,1)$ is a scaling factor. We choose $R_t=5$ and $k = 0.1$ for our experiments. $Pr(t|p)$ is taken from the partial reward matrix $M$ \footnote{Please refer to Appendix B of the supplementary for details about the construction of $M$.}, where each row has the probability distribution of the relative "closeness" of all the parent objects, to a given target object. The closeness is defined based on the relative spatial distance (measured in terms of the L2 distance) between a pair of objects in our floorplans. If multiple parent objects are $\mathtt{visible}$, we only choose the one with the maximum $R_p$. Moreover, the agent does not get this reward the next time it sees the same parent object. This encourages it to explore different parent objects in the room until the target is located. If the "$\mathtt{Done}$" action is sampled, \textit{i.e.} the termination criterion occurs, and $t$ is $\mathtt{visible}$, the agent gets the goal reward, which is the sum of $R_t$ and $R_p$. In this way, it learns to associate parent objects with a target, as well as the current state. Since the entire network is trained end-to-end, this shaped reward propagates back to the gcn layers, and tunes them to correctly learn the "parent-target" hierarchical relationship from the input knowledge graph. Finally, if neither the parent nor the target object is $\mathtt{visible}$, the agent gets a negative step penalty of $0.01$. The reward for the state $s$, and action $a$, is therefore given by:
\begin{wrapfigure}[17]{R}{0.55\textwidth} 
\begin{algorithm}[H]
\SetAlgoLined
\KwIn{state $s$, action $a$, target $t \in T$, $\mathtt{SeenList}$}
\KwData{target reward $R_t$, partial reward matrix $M$}
\SetKwFunction{FMain}{Judge}
\SetKwFunction{FMainn}{Partial}
\SetKwProg{Fn}{Function}{:}{}
\Fn{\FMain{$s$, $a$, $t$}}{
\uIf{$a\neq$ "$\mathtt{DONE}$"}{
    reward = \FMainn($s,t$)}
\uElseIf{$a == $ "$\mathtt{DONE}$" \text{\upshape and} t \text{\upshape is} $\mathtt{visible}$}{
    $\mathtt{SeenList}$ = []\;
    reward = $R_t$ + \FMainn($s,t$)\;
    }
}
\KwRet reward\;
\SetKwProg{Pn}{Function}{:}{}
\Fn{\FMainn{$s,t$}}{
    \ForEach {parent $p_i \in P$}{
        \uIf {$p_i$ is $\mathtt{visible}$ \text{\upshape and} $p_i \notin$ $\mathtt{SeenList}$}{
        $p \leftarrow argmax \text{ }M[t]$\;
        $\mathtt{SeenList}$ $\leftarrow p $\;
        $R_p$ = $M[t][p] * R_t* k$
    }
}
}
\KwRet $R_p$\;
\caption{Reward Shaping for MJOLNIR}
\label{algo}
\end{algorithm}
\end{wrapfigure}
\begin{equation*}
R(s,a) = \left\{\begin{array}{ll}
{\text{$R_p$},} & {\text {if p is $\mathtt{visible}$}}\\
{\text{$R_t$},} & {\text {if t is $\mathtt{visible}$ at termination}}\\
{\text{$R_t$}+\text{$R_p$},} & {\text {if both are $\mathtt{visible}$ at termination}}\\
{-0.01,} & {\text{otherwise}}
\end{array}\right.
\end{equation*}

Algorithm \ref{algo} summarizes the above explained process.

%% file: files/experiments.tex
\section{Experiments and Results} \label{sect:exp_res}
\subsection{Experimental setting} \label{subsect:obj_desc}
We primarily use the AI2-THOR (The House Of inteRactions) \cite{kolve2017ai2} environment as our platform for the navigation tasks. It is a challenging simulation platform, consisting of $120$ photo-realistic floorplans categorized into $4$ different room layouts - $\mathtt{Kitchen}$, $\mathtt{Living\text{ }room}$, $\mathtt{Bedroom}$, and $\mathtt{Bathroom}$. Each scene is populated with real-world objects which the agent can observe and interact with, thereby enabling algorithms trained here to be easily transferable to real-robot settings. Out of the $30$ floorplans for each scene layout, we consider the first $20$ rooms from each scene type for the training set, and the remaining $10$ rooms as the test set in our experiments. The list of target and parent objects can be found in Appendix C of the supplementary.



\subsection{Comparison Models} 
\textbf{Random} - In this model, as each step, the agent randomly samples its actions from the action space with a uniform distribution. \textbf{Baseline} - This model closely resembles that of Zhu \etal \cite{zhu2017target}, as it comprises of the current observation (in the form of the ResNet features of the current RGB frame) and the target information (in the form of glove embedding of the target object) as its state. \textbf{Scene Prior} (SP) - The publicly available implementation of Yang \etal \cite{yang2018visual} has been used here. This uses the prior knowledge in the form of a knowledge graph, but does not utilize the hierarchical relationships between objects. \textbf{SAVN} - In this model \cite{Wortsman_2019_CVPR}, the agent keeps learning about its environment through an interactive loss function even during inference time.

\textbf{Metrics} - For fair comparison with other state-of-the-art algorithms, we use the evaluation metrics proposed by \cite{anderson2018evaluation}, and adopted by other target-driven visual navigation algorithms \cite{zhu2017target, Wortsman_2019_CVPR, yang2018visual, 8963758}. The Success Rate (SR) is defined as $\frac{1}{N}\sum_{i=1}^nS_i$, while the Success weighted by Path Length (SPL) is given by $\frac{1}{N}\sum_{i=1}^nS_i\frac{l_i}{\max(l_i,e_i)}$. Here, $N$ is the number of episodes, $S_i$ is a binary vector indicating the success of the $i$-th episode. $e_i$ denotes the path length of an agent episode, and $l_i$ is the optimal trajectory length to any instance of the target object in a given scene from the initial state. We evaluate the performance of all the models on the trajectories where the optimal path length is at least $1$ ($L>=1$), and at least $5$ ($L>=5$). 

\subsection{Implementation details} 

We built our models on the publicly available code of \cite{Wortsman_2019_CVPR}, using the PyTorch framework. 
The agent was trained for $3$ million episodes on the offline data from AI2-THOR $v1.0.1$ \cite{kolve2017ai2}. During evaluation, we used $250$ episodes for each of the $4$ room types, resulting in $1000$ episodes in total. In each episode, the floorplan, target, and the initial agent position was randomly chosen from the test set defined in Section \ref{subsect:obj_desc}. Additional implementation details can be found in Appendix D of the supplementary.

\input{tables/sota_comparison}
\subsection{Results}

Table \ref{tab: model comparison} and Figure \ref{fig:test_perf} shows the performance of each of the models on the test environments. It is to be noted that the test environments consist of rooms that the agent has not previously seen during training, and therefore the location of the different objects in completely unknown. It can be seen that both of our models significantly outperform the current state-of-the-art. MJOLNIR-o has $82.9\%$ increase in SR. This supports our hypothesis that incorporating $context$ $vector$ into the observation state is indeed a better idea than directly using ResNet and GloVe features. This is because the semantic information extracted from the scene in this manner is more object-centric, thereby making the target-driven navigation problem easier. It is also important to note here, that the graph convolutional network of \cite{yang2018visual}, which does not include the $context$ $vector$ as its node feature, performs poorer than even the baseline \cite{zhu2017target} and SAVN \cite{Wortsman_2019_CVPR}. 

It is interesting to note that even though MJOLNIR-r cannot beat the performance of MJOLNIR-o, it is still a substantial improvement over the other state-of-the-art methods (an observed gain of $53.5\%$). This highlights the importance of our proposed CGN stream which can better capture the contextual information extracted via ResNet $\mathtt{conv}$ features from the current input image. Moreover, using reward shaping to tune the model parameters ensures that not only is the prior memory preserved, but the current information containing the parent-target object relationship hierarchy is also utilized.
\input{images/sr_plots}

In Figure \ref{fig:train_test_cr}, we show the convergence rate of our MJOLNIR with Yang \etal \cite{yang2018visual}. It can be seen that the training and testing SR of our algorithms rapidly increase within the first $5$ million episodes itself, before saturating. This shows that our models learn to correctly locate targets much faster than others. In contrast, for \cite{yang2018visual}, even though the training SR is quite high ($\approx 90\%$), there is huge drop during the testing performance ($\approx 35\%$), signifying severe overfitting. Finally, Table \ref{tab: room_wise} gives the comparison of room-wise results.
\input{tables/room_wise}
\subsection{Ablation study}
We show a number of ablations on MJOLNIR in Table \ref{tab: ablation}. In MJOLNIR-o (no\_g), we remove the third graph convolution layer of MJOLNIR-o, where we concatenated the intermediate embedding with the $context$ $vector$. Instead, here, we directly feed the intermediate embedding to the joint embedding. The slightly lower performance shows that the context vector can indeed help in learning meaningful node embeddings for the graph. In MJOLNIR-o (w), we used a weighted adjacency matrix for the graph convolutional network. This does not affect the performance significantly as we believe that the object-object relationship is inherently learnt by MJOLNIR-o.

\input{tables/ablation}
\vspace{-10pt}
We also evaluated our model with a different stopping criteria. In this case, the agent does not rely only on its sampled "$\mathtt{DONE}$" action to learn the termination action. Instead, it stops even when the environment gives the signal that the target object has been found. We compare our model with the SAVN \cite{Wortsman_2019_CVPR} model using this stopping criteria. The results show that we perform significantly better.

Some failure case analysis provides insights into our models, and also opens up research in further directions. We identified mainly two such cases. Firstly, the agent gets stuck at a particular state, when the target is visible, but the straight path to it blocked by some obstacle. We hypothesize that a planner module which checks for collision might help to overcome this. Secondly, since "$\mathtt{DONE}$" is one of the candidate actions, (i) it may be wrongly sampled even when the goal hasn't been reached, or (ii) it might not be sampled even after reaching the goal. A workaround is to have the termination criterion provided by the environment. As seen from our ablation study, this boosts the success rate from $65.3\%$ to $83.1\%$. However, it is at the cost of increased episode length as the agent is encouraged to explore more of the environment. Moreover, in a real-robot setting, it might not be possible to have the environment signal the termination.

%% file: tables/sota_comparison.tex
\begin{table}[t]
\begin{center}
\small
\ra{1.2}
\scalebox{1}{
\begin{tabular}{lccccc}\toprule
 & \multicolumn{2}{c}{$L>=1$} && \multicolumn{2}{c}{$L>=5$} \\
\cline{2-3} \cline{5-6}
Model & SR(\%) & SPL(\%) && SR(\%) & SPL(\%) \\
\hline
Random &  11.2 & 5.1 && 1.1 & 0.50\\
Baseline \cite{zhu2017target} & 35.0 & 10.3 && 25.0 & 10.5\\
Scene-prior \cite{yang2018visual} & 35.4 & 10.9 && 23.8 & 10.7\\
SAVN \cite{Wortsman_2019_CVPR}& 35.7 & 9.3 && 23.9 & 9.4\\
\hline
\textbf{MJOLNIR-r (our)}& 54.8 & 19.2 && 41.7 & 18.9\\
\textbf{MJOLNIR-o (our)}& \textbf{65.3} & \textbf{21.1} && \textbf{50.0} & \textbf{20.9}\\
\bottomrule
\end{tabular}
}
\end{center}
\caption{Comparison with state-of-the-art visual navigation algorithms on the unseen test set}
\label{tab: model comparison}
\end{table}

%% file: images/sr_plots.tex
\begin{figure}[h]
    \centering
    \begin{subfigure}[h]{0.49\linewidth}
        \centering
        \includegraphics[height=1.8in]{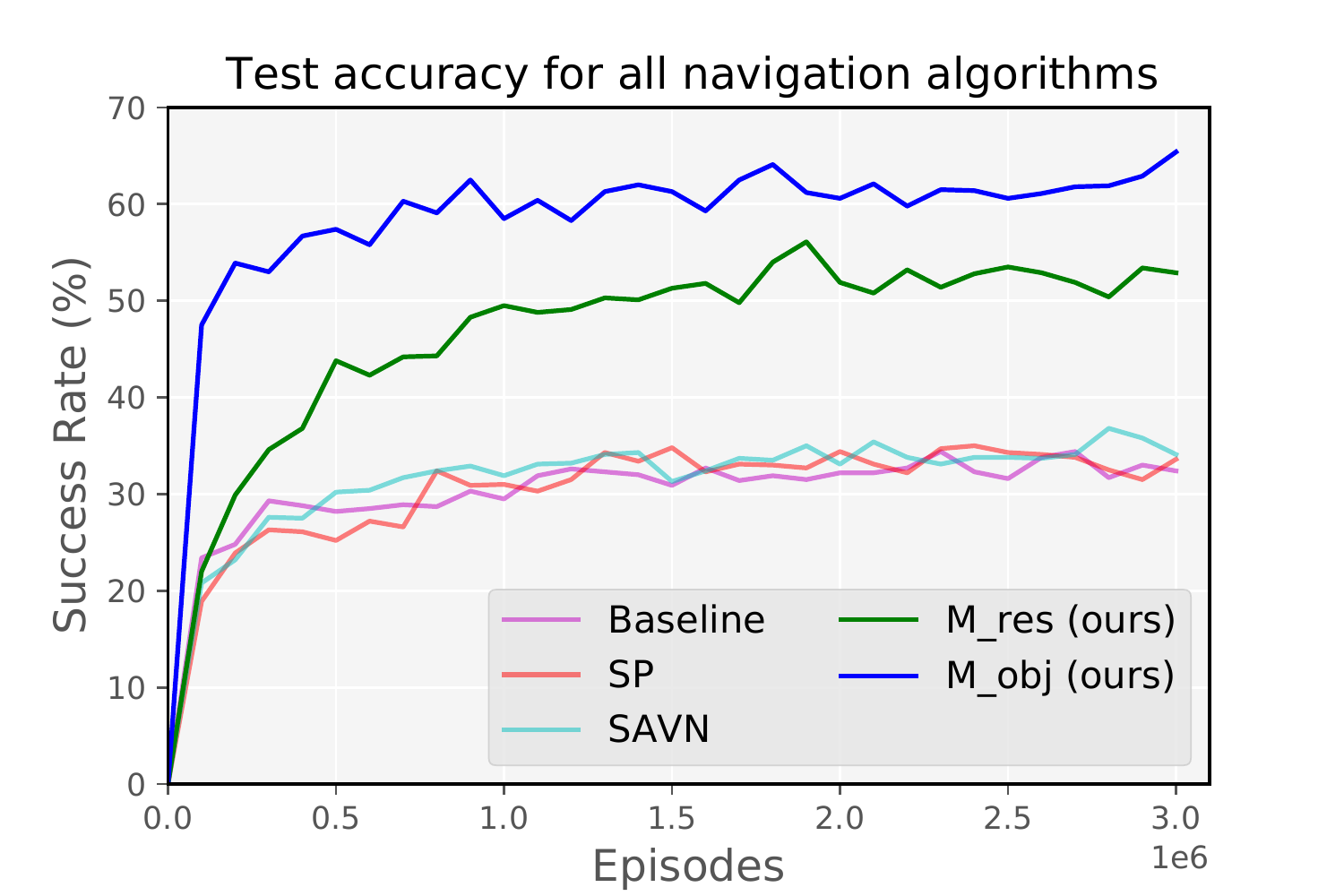}
        \caption{Test accuracy for all models}
        \label{fig:test_perf}
    \end{subfigure}
    ~ 
    \begin{subfigure}[h]{0.49\linewidth}
        \centering
        \includegraphics[height=1.8in]{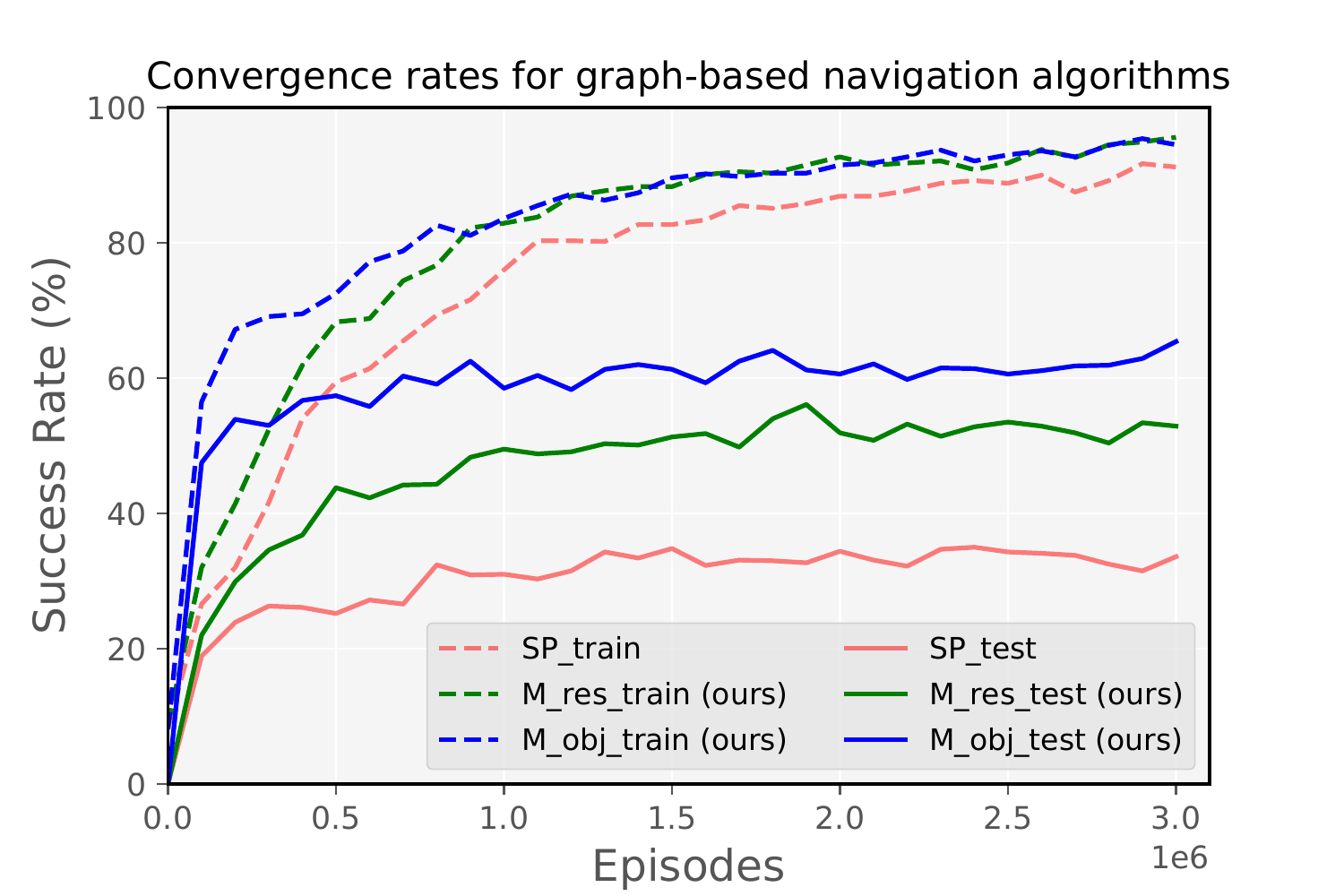}
        \caption{Convergence rate for graph-based models}
        \label{fig:train_test_cr}
    \end{subfigure}%
    \caption{Test accuracy and convergence rates for all the algorithms}
\end{figure}

%% file: tables/room_wise.tex
\begin{table}[t]
\begin{center}
\small
\ra{1.2}
\scalebox{0.83}{
\begin{tabular}{|l|cc|cc|cc|cc|cc|}
\hline
\multicolumn{1}{|c|}{\multirow{2}{*}{Models}} & \multicolumn{2}{c|}{Bathroom} & \multicolumn{2}{c|}{Bedroom} & \multicolumn{2}{c|}{Kitchen} & \multicolumn{2}{c|}{Living Room} & \multicolumn{2}{c|}{Average}\\
 & SR(\%) & SPL(\%) & SR(\%) & SPL(\%) & SR(\%) & SPL(\%) & SR(\%) & SPL(\%) & SR(\%) & SPL(\%)\\
\hline
Baseline \cite{zhu2017target} & 53.2 & 13.4 &28.8& 9.0 &32.4&10.9&35.2&10.0&37.4&10.8 \\
Scene Prior \cite{yang2018visual} & 41.6 & 13.3 & 33.6 & 10.4 & 26.4 & 9.1 & 36.0 & 9.9 & 34.4 & 10.7\\
SAVN \cite{Wortsman_2019_CVPR}& 47.6 & 14.6 & 21.6 & 6.7 & 34.8 & 8.3 & 40.0 & 9.0 & 36.9 & 9.7\\
\hline
\textbf{MJOLNIR-r (our)} & 72.8 & 24.3 & 41.2 & \textbf{16.9} & 56.4 & 21.2 & \textbf{50.8} & 15.9 & 55.3 & 19.6\\
\textbf{MJOLNIR-o (our)} & \textbf{82.4} & \textbf{25.1} & \textbf{43.2} & \text{14.4} & \textbf{74.8} & \textbf{22.9} & 50.0 & \textbf{17.9} & \textbf{62.6} & \textbf{20.1}\\
\hline
\end{tabular}
}
\end{center}
\caption{Evaluation results on a per-room basis}
\label{tab: room_wise}
\end{table}

%% file: tables/ablation.tex
\begin{table}[h]
\begin{center}
\small
\ra{1.2}
\scalebox{1}{
\begin{tabular}{|l|l|ccccc|}\hline
\multirow{2}{*}{\shortstack[l]{$\mathtt{DONE}$ \\ action}}  & \multicolumn{1}{c|}{\multirow{2}{*}{Model}} & \multicolumn{2}{c}{$L>=1$} && \multicolumn{2}{c|}{$L>=5$} \\
\cline{3-4} \cline{6-7}
 && SR(\%) & SPL(\%) && SR(\%) & SPL(\%) \\
\hline
\multirow{4}{*}{\shortstack[l]{only \\ sampled}}
& MJOLNIR-r & 54.8 & 19.2 && 41.7 & 18.9\\
& MJOLNIR-o  & \textbf{65.3} & 21.1 && \textbf{50.0} & \textbf{20.9}\\
& MJOLNIR-o (no\_g) & 59.0 & 16.6 && 41.0 & 16.9\\
& MJOLNIR-o (w) & 64.7 & \textbf{21.6} && 46.4 & 20.6\\
\hline \hline
\multirow{2}{*}{\shortstack[l]{sampled\\ $+$ env}}
& SAVN \cite{Wortsman_2019_CVPR} & 54.4 & 35.55 && 37.87 & 23.47\\
& MJOLNIR-o & \textbf{83.1} & \textbf{53.9} && \textbf{71.6} & \textbf{36.9}\\
\hline
\end{tabular}
}
\end{center}
\caption{Ablation study for MJOLNIR}
\label{tab: ablation}
\end{table}

%% file: files/conclusion.tex
\section{Conclusion} \label{sect:conc}
We introduced MJOLNIR, a novel object-goal visual navigation algorithm that utilizes prior knowledge, and also learns to associate object "closeness" in the form of parent-target hierarchy during training. This is done through the proposed {\em context vector} which can be easily derived from the output of an object detector. We show that besides the modified state space, knowledge graph and reward shaping too play a significant role in guiding the agent to search for the target. Extensive experiments show that the agent can successfully find small target objects using the larger parent object as an anchor. Our model has the ability to generalize across different unseen scenes, and current state-of-the-art models.


%% file: files/appendix.tex
\newpage
\begin{appendices}
\section*{Appendix A: Object detector vs ResNet features} \label{obj_vs_res}
One of the significant changes made in our algorithm from existing works \cite{yang2018visual} is to use object detection features instead of a classification probability from a CNN such as ResNet as the input node embedding in the graph convolution network. This is because detection of multiple objects is integral to learning the "parent-target" hierarchical object relationships effectively. To showcase this, we run a YOLOv3 detection, $\mathtt{yolo.mp4}$, along with a Grad-cam visualization of ResNet-18, $\mathtt{resnet.mp4}$, on a cell phone recording of an indoor kitchen scene. As seen from the videos, the multi-object detector can generate much less false-detections as compared to the 1000 class probability from ResNet, which generates noisy labels such as "$\mathtt{prison}$" and "$\mathtt{home\_theatre}$".
\input{tables/m_tables}
\section*{Appendix B: Construction of Partial reward matrix $M$} \label{m_matrix}
As mentioned in Section 4 of the main paper, the conditional probability of a target being found, given that a parents was observed, $Pr(t|p)$, is obtained from the Partial reward matrix $M$. We utilized the training split of our environment, AI2-THOR, for creating $M$. For every floorplan, the 3D position of each object present there was plotted. Then, we counted the occurrence of different target objects which were located within a euclidean distance of 1 meter from a parent object (from Table \ref{tab: par_tar_list}). It is to be noted that even though the same parent object might be present in more than one room type, its relationship with the target objects might be different. Thus, we computed $M$ for every room type, as shown in Table \ref{tab:full_m}. Normalizing each row provides the probability distribution of a target object $t \in T$, given a parent object $Pr(t|P)$. For the final parent reward $R_p$, we also used a scaling factor, $k = 0.1$ to ensure that the agent receives lesser reward for the parent object as compared to the target reward $R_t=5$.
\section*{Appendix C: Parent and target object list} \label{par_tar_list}
\begin{table}[!h]
\begin{center}
\ra{1.2}
\scalebox{0.64}{
\begin{tabular}{|l|c|c|}
\hline
\textbf{Room type} & \textbf{Target Objects} $T$ & \textbf{Parent Objects} $P$\\
\hline
Kitchen & Toaster, Spatula, Bread, Mug, CoffeeMachine, Apple & Fridge, StoveBurner, Microwave, TableTop, Sink, CounterTop, Shelf\\
Living room & Painting, Laptop, Television, RemoteControl, Vase, ArmChair & Drawer, Shelf, TableTop, Sofa, FloorLamp\\
Bedroom & Blinds, DeskLamp, Pillow, AlarmClock, CD & Shelf, Dresser, NightStand, Drawer, Desk, Bed\\
Bathroom & Mirror, ToiletPaper, SoapBar, Towel, SprayBottle & CounterTop, Cabinet,  Drawer,  ShowerDoor, Toilet, Bathtub\\
\hline
\end{tabular}}
\end{center}
\caption{Parent and target object list}
\label{tab: par_tar_list}
\end{table}

Table \ref{tab: par_tar_list} provides the list of target objects, $T$, and parent objects, $P$, used in our experiments. 

\section*{Appendix D: Implementation Details} \label{impl_det}
For the word embeddings, we used the $300$-D GloVe vectors pretrained on $840$ billion tokens of Common Crawl \cite{pennington2014glove}. The A3C model is based on \cite{pytorchaaac}. 
The model hyperparameters used for our experiments are tabulated below.

\begin{table}[!h]
\begin{center}
\ra{1.2}
\scalebox{0.8}{
\begin{tabular}{|l|c|c|c|c|c|}
\hline
\textbf{Parameters} & \textbf{Baseline} \cite{zhu2017target} & \textbf{Scene Prior} \cite{yang2018visual} & \textbf{SAVN} \cite{Wortsman_2019_CVPR} & \textbf{MJOLNIR-r (our)} & \textbf{MJOLNIR-o (our)} \\
\hline
Learning rate & 0.0001 & 0.0001 & 0.0001 & 0.0001 & 0.0001\\
Optimizer & SharedAdam & SharedAdam & SharedAdam & SharedAdam & SharedAdam\\
Discount Factor & 0.99 & 0.99 & 0.99 & 0.99 & 0.99\\
max \#workers & 8 & 8 & 6 & 8 & 8\\
\multirow{2}{*}{}Observation stream & ResNet & ResNet & ResNet & ResNet & Object detector \\ encoder & & & & &\\
Max training episodes & $3\times10^6$ & $3\times10^6$ & $3\times10^6$ & $3\times10^6$ & $3\times10^6$\\
\hline
\end{tabular}}
\end{center}
\caption{Summary of Hyperparameters}
\label{tab: parameters}
\end{table}
\end{appendices}

%% file: tables/m_tables.tex
\begin{table}[!h]
    \begin{subtable}{\linewidth}
      \centering
    \ra{1.2}
    \scalebox{0.9}{
    \begin{tabular}{|l|c|c|c|c|c|c|c|}
        \hline
        \backslashbox{Target}{Parent} & Fridge &  StoveBurner &  Microwave &  TableTop &  Sink &  CounterTop & Shelf \\
        \hline
        Toaster & 0.15 & \textbf{0.29} & 0.15 & 0.04 & 0.15 & 0.23 & - \\
        Spatula & 0.03 & \textbf{0.31} & 0.22 & 0.02 & 0.19 & 0.22 &  0.02 \\
        Bread & - & 0.16 & 0.13 & 0.16 & 0.20 & \textbf{0.36} & - \\
        Mug & - & 0.19 & 0.17 & 0.11 & \textbf{0.30} & 0.23 & - \\
        CoffeeMachine & 0.08 & 0.10 & 0.10 & 0.06 & \textbf{0.38} & 0.28 & - \\
        Apple & 0.12 & 0.12 & 0.12 & 0.12 & \textbf{0.25} & 0.23 & 0.02 \\
        \hline
    \end{tabular}}
    \caption{Kitchen}
    \label{tab:Kitchen}
    \end{subtable}%
    \\ \\ \\
    \begin{subtable}{.5\linewidth}
      \centering
    \ra{1.2}
    \scalebox{0.65}{
    \begin{tabular}{|l|c|c|c|c|c|}
    \hline
    \backslashbox{Target}{Parent} &  Drawer & Shelf & TableTop &  Sofa & FloorLamp \\
    \hline
    Painting      &    \textbf{0.86} &  0.14 &        - &     - &         - \\
    Laptop        &    0.27 &  0.14 &     \textbf{0.36} &  0.23 &         - \\
    Television    &    \textbf{0.40} &   0.2 &     0.35 &     - &      0.05 \\
    RemoteControl &    \textbf{0.25} &  0.05 &     \textbf{0.25} &   0.4 &      0.05 \\
    Vase          &    0.21 &  \textbf{0.47} &     0.26 &     - &      0.05 \\
    ArmChair      &    0.09 &     - &     0.27 &  0.18 &      \textbf{0.45} \\
    \hline
    \end{tabular}}
    \caption{Living room}
    \label{tab:livingroom}
    \end{subtable}%
    \begin{subtable}{.5\linewidth}
      \centering
    \ra{1.2}
    \scalebox{0.7}{
    \begin{tabular}{|l|c|c|c|c|c|c|}
    \hline
    \backslashbox{Target}{Parent} & Shelf & Dresser & NightStand & Drawer &  Desk &   Bed \\
    \hline
    Blinds     &     \textbf{1} &       - &          - &      - &     - &     - \\
    DeskLamp   &  0.16 &     0.2 &       0.12 &   0.24 &  \textbf{0.28} &     - \\
    Pillow     &     - &    0.04 &       0.21 &   0.17 &     - &  \textbf{0.58} \\
    AlarmClock &  0.19 &    0.11 &       \textbf{0.26} &   0.21 &  0.04 &  0.19 \\
    CD         &  0.21 &     0.1 &       0.08 &   \textbf{0.33} &  0.23 &  0.06 \\
    \hline
    \end{tabular}}
    \caption{Bedroom}
    \label{tab:Bedroom}
    \end{subtable}%
    \\ \\ \\
    \begin{subtable}{\linewidth}
      \centering
    \ra{1.2}
    \scalebox{1}{
    \begin{tabular}{|l|c|c|c|c|c|c|}
    \hline
    \backslashbox{Target}{Parent} &  CounterTop &  Cabinet &  Drawer &  ShowerDoor & Toilet & Bathtub \\
    \hline
    Mirror      &        \textbf{0.51} &     0.26 &    0.21 &        0.03 &      - &       - \\
    ToiletPaper &        0.19 &     0.19 &    0.10 &        0.06 &   \textbf{0.46} &       - \\
    SoapBar     &        \textbf{0.29} &     0.18 &    0.14 &        0.04 &    0.2 &    0.16 \\
    Towel       &        0.15 &     0.04 &    0.15 &        \textbf{0.44} &   0.07 &    0.15 \\
    SprayBottle &        \textbf{0.35} &     0.20 &    0.16 &        0.02 &   0.22 &    0.06 \\
    \hline
    \end{tabular}}
    \caption{Bathroom}
    \label{tab:Bathroom}
    \end{subtable}%
    \caption{Partial reward matrices for each of the 4 room-type in AI2-THOR.}
    \label{tab:full_m}
\end{table}